\documentclass[10pt,twocolumn,letterpaper]{article}

\usepackage{3dv}
\usepackage{times}
\usepackage{epsfig}
\usepackage{graphicx}
\usepackage{amsmath}
\usepackage{amssymb}
\usepackage{subfig}

\usepackage{multirow}

\usepackage[pagebackref=true,breaklinks=true,letterpaper=true,colorlinks,bookmarks=false]{hyperref}

\threedvfinalcopy 

\def\threedvPaperID{327} 

\ifthreedvfinal\fi
\begin{document}

\title{A New Distributional Ranking Loss With Uncertainty: Illustrated in \\Relative Depth Estimation}

\author{
Alican Mertan, Yusuf Huseyin Sahin, Damien Jade Duff, and Gozde Unal\\
Istanbul Technical University, Istanbul, Turkey\\
{\tt\small \{mertana, sahinyu, djduff, gozde.unal\}@itu.edu.tr}
}

\maketitle

\begin{abstract}
   We propose a new approach for the problem of relative depth estimation from a single image. Instead of directly regressing over depth scores, we formulate the problem as estimation of a probability distribution over depth and aim to learn the parameters of the distributions which maximize the likelihood of the given data. To train our model, we propose a new ranking loss, Distributional Loss, which tries to increase the probability of farther pixel's depth being greater than the closer pixel's depth. Our proposed approach allows our model to output confidence in its estimation in the form of standard deviation of the distribution. We achieve state of the art results against a number of baselines while providing confidence in our estimations. Our analysis show that estimated confidence is actually a good indicator of accuracy. We investigate the usage of confidence information in a downstream task of metric depth estimation, to increase its performance.
\end{abstract}

\section{Introduction}


Depth is a key factor of a scene and it has always been an important challenge to estimate it, especially from monocular images. With the advancements in the deep learning techniques, we started to see very successful attempts in monocular depth estimation task such as \cite{eigen_depth_2014, laina_deeper_2016, fu_deep_2018}, where the aim is to estimate absolute depth. However, most of the state of the art works focusing on absolute depth estimation, utilize limited datasets such as indoor only (\eg NYUDv2) or outdoor only (\eg KITTI) datasets. While models trained on limited datasets perform well on their immediate training domain, they do not generalize well to images coming from different distributions.

In order to be able to estimate depth in-the-wild, a reformulation of the depth estimation problem is employed, namely relative depth estimation. With this reformulation, diverse datasets are collected and models that work in-the-wild are trained \cite{chen_single-image_2016, xian_monocular_2018, chen_learning_2019}. However, these approaches are not capable of expressing model confidence, which we believe to be a very important information for the depth estimation problem, as the  estimated depth map is usually utilized for subsequent tasks or decision making processes. For instance, both \cite{pizzoli_remode_2014, tateno_cnn-slam_2017} utilize uncertainty maps in 3D reconstruction to increase performance and robustness.

\begin{figure}
     \centering
     \subfloat[][RGB]{\includegraphics[width=0.15\textwidth]{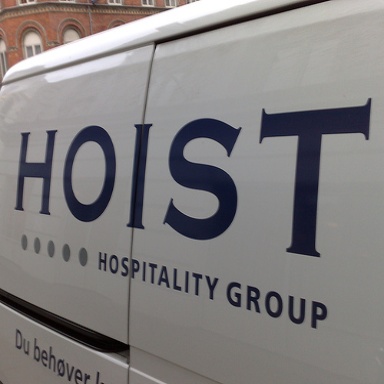}
     \label{fig:rgb}}
     \subfloat[][Depth]{\includegraphics[width=0.15\textwidth]{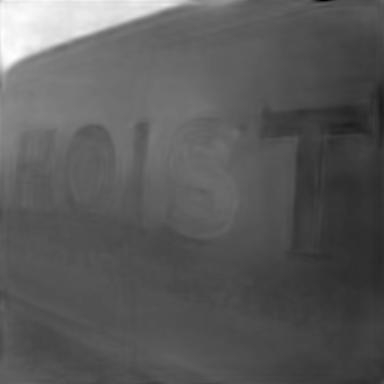}
     \label{fig:depth}}
     \subfloat[][Confidence]{\includegraphics[width=0.15\textwidth]{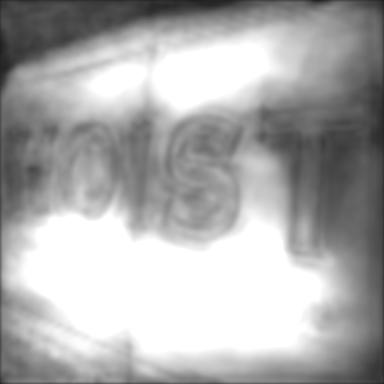}
     \label{fig:confidence}}\vspace{-0.2cm}
     \caption{A model trained with the proposed Distributional ranking loss learns to output a dense confidence map for its predictions. Fig.~\ref{fig:rgb} shows the input RGB image where we see letters on the side of a vehicle. However in Fig. \ref{fig:depth}, it can be seen that the model fails to account for the smooth depth change across the letters, and produces unreliable depth estimates that normally should overlook the letter boundaries. While this mistake would go unnoticed for previous approaches, our model learns to express its estimation confidence. Fig.~\ref{fig:confidence} clearly shows that the model is not confident in its estimations for the letters, indicated by darker pixels. } 
     \label{fig:intro_example}
\end{figure}

While previous approaches directly estimate depth scores as a regression task, we treat depth as it is normally distributed, parameterized by mean $\mu$ and standard deviation $\sigma$, and regress over these parameters for each pixel. We believe that this representation is more natural as the ground truth information is also uncertain about the actual depth value. Furthermore, this representation effectively makes our model capable of displaying its confidence in terms of $\sigma$. In order to learn from ordinal relations of pixels, we propose a novel loss function, a distributional loss, which attempts to increase the probability of farther pixel's depth being greater than closer pixel's depth. An illustrative example is shown in Figure~\ref{fig:intro_example}, where our model outputs a depth map as well as a separate confidence map for the estimated depth, which points to regions of uncertainty in the estimation.

The contributions of our work are as follows:
\begin{itemize}
    \item We formulate the problem of depth estimation as estimating a probability distribution over depth where $\frac{1}{\sigma}$ can be considered as the confidence. Given the ground truth information, we believe this formulation is more intuitive and it allows us to output confidence.
    \item We devise a new ranking loss, the distributional loss, that allows us to learn parameters of the distribution for each pixel from ordinal relations of pixels.
    \item We evaluate our approach against a number of baselines in the literature and achieve state of the art performances.
    \item We analyze the confidence output and empirically exhibit its usefulness.
\end{itemize}

\section{Background}
\textbf{Absolute depth estimation} Early works in the field utilized hand crafted features and Markov Random Fields while incorporating human expertise in terms of hand designed constraints on optimization process \cite{hoiem_automatic_2005, michels_high_2005, saxena_learning_2006, saxena_make3d:_2008, liu_single_2010}. With the increasing success of convolutional neural networks on vision problems, a number of works employed convolutional neural networks in a standard supervised learning setting \cite{eigen_depth_2014, laina_deeper_2016, cao_estimating_2017, fu_deep_2018}. Additional complementary tasks were also utilized to increase the performance \cite{eigen_predicting_2015, wang_towards_2015, qi_geonet_2018, xu_pad-net_2018, zhang_joint_2018, zhang_pattern-affinitive_2019, chen_towards_2019}. In order to eliminate the need for real world ground truth data, number of works do self-supervised learning \cite{garg_unsupervised_2016, godard_unsupervised_2017, zhou_unsupervised_2017, wang_learning_2018, ranjan_competitive_2018}, while \cite{ren_cross-domain_2018} used synthetic images. 

\textbf{Relative depth estimation} To the best of our knowledge, Zoran \etal \cite{zoran_learning_2015} did the first attempt at relative depth estimation by classifying ordinal relations of pixel pairs. Since it is infeasible to classify all possible pixel pairs, they superpixelated the input image and only compared centers of superpixels, assuming that superpixels represent homogeneous depth patches. \cite{chen_single-image_2016, chen_learning_2019} estimated a dense score map in a regression setting and used a pairwise ranking loss to learn from ordinal relations. In the same framework, \cite{xian_monocular_2018} applied an improved pairwise ranking loss which focuses on a set of hard pairs and \cite{xian2020structure} proposed a sampling strategy that focuses on image and object edges. \cite{mertan_listwise} employed a listwise ranking loss which allowed their model to focus more on closer pixels.

\textbf{In-the-wild datasets} First dataset with relative depth annotations is Depth in the Wild (DIW) \cite{chen_single-image_2016}. It consists of randomly sampled images from internet and ground truth ordinal relation of one pair of pixels per image, annotated by human annotators, and has an official train test split. Afterwards, two other datasets with relative depth annotations are proposed: YouTube3D \cite{chen_learning_2019} and Relative Depth from Web (RedWeb) \cite{xian_monocular_2018}. While YouTube3D offers sparse ground truth information, RedWeb dataset has dense relative depth annotation that can be acquired from given ground truth score map. YouTube3D and RedWeb do not have an official train test split. Works that use these datasets, use the whole dataset for training and test their performance on DIW test split.

\textbf{Ranking} Ranking methods can be divided into two main categories based on whether they directly optimize ranking measures. \cite{cakir_deep_2019, revaud_learning_2019} proposes differentiable approximations for ranking measures which allow them to be used in the optimization process. On the other hand, a number of works optimize surrogate measures in a pairwise \cite{burges_learning_2005, tsai_frank:_2007} or a listwise manner \cite{cao_learning_2007, qin_query-level_2008, xia_listwise_2008, lan_position-aware_2014}. Yet none of them learns to output confidence.

\section{Approach}
In relative depth estimation, the ground truth information consists of pixels' ordinal relations which falls in three categories as
\begin{equation}
    \label{eq:gt}
    \forall i,j \in \mathbf{\Omega}, 
    \begin{cases}
    r_{ij} = 1,& \text{if } d_i > d_j\\
    r_{ij} = -1,& \text{if } d_i < d_j\\
    r_{ij} = 0,& \text{if } d_i = d_j
    \end{cases}
\end{equation}
where $\mathbf{\Omega}$ represents the set of all pixels, $d_i$ is the metric depth of pixel $i$, and $r_{ij}$ represents the ordinal relations of pixel $i$ and $j$. The aim is to predict the ordinal relation for pixel pair $i$ and $j$, $\tilde{r}_{ij}$, to minimize the following objective:
\begin{equation}
    \label{eq:satisfy_general}
    \min{\sum_{i,j \in \Omega} \mathbf{1}\left( r_{ij} \neq \tilde{r}_{ij} \right)}
\end{equation}
where $ \mathbf{1}\left(\right) $ is the indicator function that evaluates to $1$ if $r_{ij} \neq \tilde{r}_{ij}$ is true, otherwise it evaluates to $0$.

\begin{figure*}
\begin{center}
\includegraphics[width=0.9\linewidth]{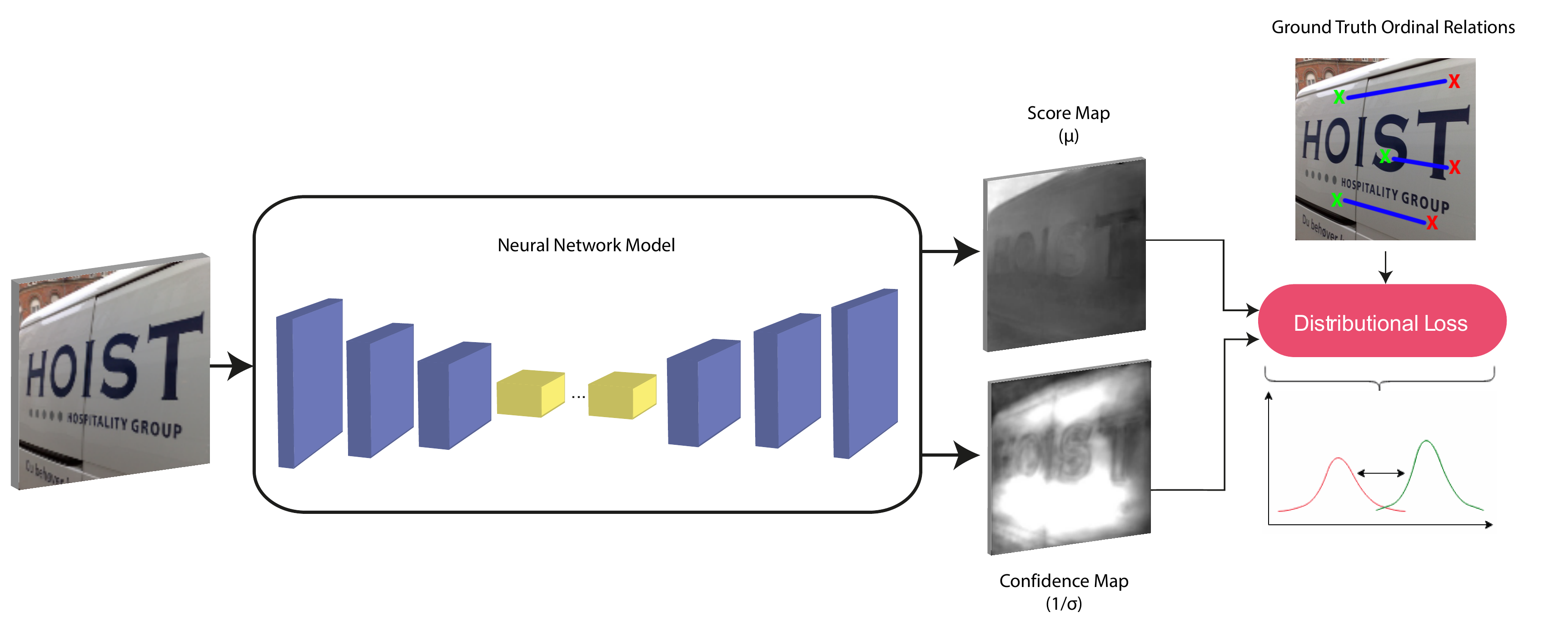}\vspace{-0.4cm}
\end{center}
\caption{Depiction of our framework. A network model regresses over $\mu$ and reciprocal of $\sigma$, $\frac{1}{\sigma}$, which represent score map and confidence map, respectively. Darker pixels show closer points in the score map and lower confidence in confidence map.}
\label{fig:framework}
\end{figure*}

We explain our approach to solve relative depth estimation problem in two main parts. First, we discuss our formulation of the relative depth estimation problem. Next, we present the ranking loss that works with the proposed formulation. Figure  \ref{fig:framework} depicts the overall framework that we use to train neural networks for relative depth estimation problem. Note that our approach does not depend on any particular neural network architecture, or a particular model for that matter. Any parameterized, differentiable model can be used as a ranking function. Additionally, we propose a general approach, in a sense that it is applicable to any other ranking problems. Yet throughout the paper, we are going to discuss our approach particularly for the relative depth estimation problem as we showcase our approach in this domain. 

\subsection{Formulation}
As previous works \cite{chen_single-image_2016, xian_monocular_2018, chen_learning_2019}, we follow a pairwise approach. We consider two relations between pixels where either pixel $i$ is farther compared to pixel $j$, or pixel $j$ is farther compared to pixel $i$. We omit the case where both pixels are at the same depth as DIW and YouTube3D datasets do not have examples of equality case.

To solve the relative depth estimation problem, previous works try to estimate a score for each pixel that can be used as ordinal relations as follows:
\begin{equation}
    \label{eq:scalar_form}
    \forall i,j \in \mathbf{\Omega}, 
    \begin{cases}
    \tilde{r}_{ij} = 1, & \text{if } \tilde{s}_i > \tilde{s}_j  \\
    \tilde{r}_{ij} = -1, & \text{if } \tilde{s}_i < \tilde{s}_j  
    \end{cases}
\end{equation}
where $\tilde{s}_i$ is the estimated score for pixel $i$. 

However, we neither know the ground truth values for $\tilde{s}_i,~\forall i$, nor do care about actual value of $\tilde{s}_i,~\forall i$, as long as they satisfy Equation~(\ref{eq:scalar_form}). In these circumstances, we believe it is more intuitive to estimate a probability distribution over depth values, rather than scalar values. In this work, we investigate the employment of one natural choice for this task, that is the  normal distribution. 

Probability of depth of pixel $i$ conditional on an image $I$, is modelled as a normal distribution whose mean $\mu_i$ and standard deviation $\sigma_i$ is estimated from the image $I$ by a neural network model $g(I;\theta)$:
\begin{equation}
    \label{eq:p_over_d}
    \begin{gathered}
        P(d_i | I) \sim \mathcal{N}( \mu_i, \sigma_i )\\
        \text{where } ~\mu_i, \sigma_i = g(I;\theta).
    \end{gathered}
\end{equation}

We formulate the relative depth estimation problem as a maximum likelihood estimation problem and try to find the optimal parameters $\hat{\theta}$ that maximizes the probability of observed data $\mathbf{y}$ as follows:
\begin{equation}
    \label{eq:MLE}
    \hat{\theta}=\underset{\theta \in \Theta}{\arg \max } ~h(\mathbf{y};\theta )
\end{equation}
where $\Theta$ represents the parameter space, and $h$ is the data likelihood function, which is to-be-defined (Eq.~\ref{eq:aim}).

With this formulation, we can consider $\sigma_i$ as an uncertainty information for that pixel $i$'s depth estimation. As $\sigma_i$ increases, the range of values, into which the pixel $i$'s score value falls, increases. 
In practice, we model the function $g(I;\theta)$ with a neural network model. When we first initialize the network, we observe small values in the outputs, indicating that there is very small uncertainty in model's estimations. This is counter intuitive, since at the start of the training we expect the uncertainty to start at high levels and decrease as the training proceeds. Also, estimating $\sigma_i$ is not numerically stable for the very same reason. To overcome this problem, we propose to learn to estimate the reciprocal of $\sigma_i$, i.e. $\frac{1}{\sigma_i}$, which can be interpreted as the "confidence". This formulation reflects the model's knowledge in a better way, particularly addressing the low confidence at the start of the training. We also experiment with both of the formulations and empirically show that learning the reciprocal, \ie the confidence rather than the uncertainty, is better.

\subsection{Distributional Loss (DL)}
In this section, we introduce our proposed ranking loss, the distributional loss. It works in a pairwise fashion. Since ground truth information consists of ordinal relations of pixels, we calculate the likelihood of the observed training data as follows:
\begin{equation}
    \label{eq:aim}
    \begin{gathered}
        h(y;\theta) = \prod_{i,j \in \Omega} 
        \bigl(\mathbf{1}(r_{ij}=1) P(d_i > d_j) + \\ 
        \mathbf{1}(r_{ij}=-1) P(d_j > d_i)\bigr) .
    \end{gathered}
\end{equation} 
where $P$ refers to the probability of depth of pixel $i$ being greater than that of pixel $j$ or vice versa depending on its arguments, and $\mathbf{1}(\cdot)$ refers to the indicator function.

To simplify things, let us assume
\begin{equation}
    \label{eq:simplify}
    \forall i,j \in \mathbf{\Omega}, 
    \begin{cases}
    i \triangleq f,~ j \triangleq c,& \text{if } d_i > d_j\\
    j \triangleq f,~ i \triangleq c,& \text{if } d_j > d_i
    \end{cases}
\end{equation}
where $f$ refers to the pixel that is supposed to be farther and $c$ refers to the pixel that is supposed to be closer for a particular pairwise relation.
Equation~(\ref{eq:aim}) now becomes 
\begin{equation}
    \label{eq:aim_simple}
    h(y;\theta) = \prod_{f,~c ~\in ~\Omega} P(d_f>d_c),
\end{equation} 
which can be maximized by minimizing the following distributional loss
\begin{equation}
    \label{eq:DL}
    \operatorname{DL} = - \log\left( \prod_{f,~c ~\in ~\Omega} P(d_f>d_c)\right).
\end{equation}

By rearranging the terms of $P(d_f > d_c)$, we get 
\begin{equation}
\label{eq:diff_normal}
\begin{aligned}
    P(d_f > d_c)    &= P(d_f-d_c > 0)\\
                    &= P( z > 0 ), \\
    \text{where } z &\sim \mathcal{N}(\mu_z, \sigma_z^2),\\
     \mu_z \triangleq \mu_f - \mu_c&, ~\sigma_z^2 \triangleq  \sigma_f^2 + \sigma_c^2, 
\end{aligned}
\end{equation}
which can be calculated as
\begin{equation}
\label{eq:prob_Q}
    \begin{split}
    P\left(z>0\right) &= Q\left( \frac{-\mu_z}{\sigma_z} \right) \text{ or}\\
    &= \frac{1}{2} \left(1 - \Phi\left( \frac{ \frac{-\mu_z}{\sigma_z} }{\sqrt{2}} \right)\right),
    \end{split}
\end{equation}
where $\Phi$ is the error function $\operatorname{erf}(\cdot)$ in mathematics that has differentiable implementations in popular libraries. Overall, our distributional loss in its full form is as follows:
\begin{equation}
    \label{eq:dist_loss}
    \operatorname{DL} = \sum_{f,c \in \Omega} - \log\left( \frac{1}{2} \left(1 - \Phi\left( \frac{-(\mu_F-\mu_C)}{ \sqrt{2}\sqrt{\sigma_F^2 + \sigma_C^2}} \right)\right) \right).
\end{equation}

In test time, we choose $\tilde{r}_{ij}$ as follows:
\begin{equation}
    \label{eq:prob_form}
    \forall i,j \in \mathbf{\Omega}, 
    \begin{cases}
    \tilde{r}_{ij} = 1, & \text{if } ~P(d_i>d_j) > 0.5  \\
    \tilde{r}_{ij} = -1, & \text{if } ~P(d_i>d_j) < 0.5.
    \end{cases}
\end{equation}
In practice, we only compare $\mu_i$ and $\mu_j$ as $P(d_i>d_j)>0.5$ only when $\mu_i>\mu_j$ or vice versa. 

\begin{figure}
\begin{center}
\includegraphics[width=0.8\linewidth]{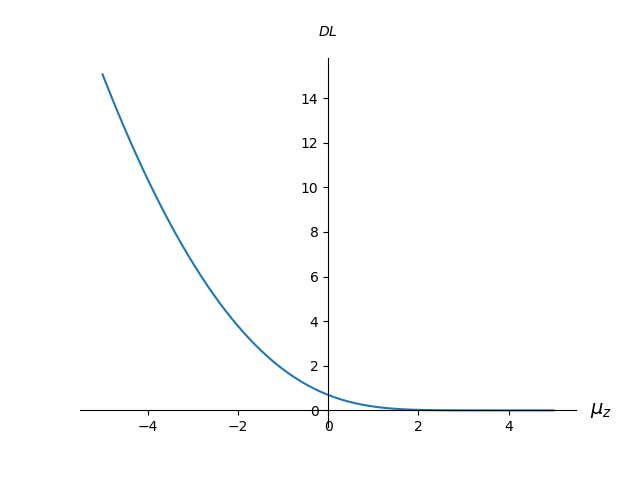}
\end{center}
\caption{DL loss vs $\mu_f-\mu_c$ where $\sqrt{\sigma_f^2+\sigma_c^2}$ is fixed to $1$.}
\label{fig:fix-sigma-var-mean}
\end{figure}

\textbf{Behaviour of the DL loss function} Let us examine how the proposed DL loss function behaves under certain conditions. First we fix the standard deviation $\sqrt{\sigma_f^2+\sigma_c^2}$ to 1 and plot the change in the loss as $\mu_f-\mu_c$ changes. As it can be seen from Figure~\ref{fig:fix-sigma-var-mean}, the DL loss decreases monotonously as the $\mu_f-\mu_c$ increases. Similar to the ranking losses used in previous works \cite{chen_single-image_2016, xian_monocular_2018, chen_learning_2019}, the distributional loss also encourages bigger differences between scores, but only to some extent. The loss vanishes as $\mu_z$ increases since as long as $P(d_f>d_c)>0.5$ is satisfied, we gain no extra benefit in terms of our objective Equation~(\ref{eq:satisfy_general}).

\begin{figure}
     \centering
     \subfloat[][$\mu_f-\mu_c = 1$]{\includegraphics[width=0.25\textwidth]{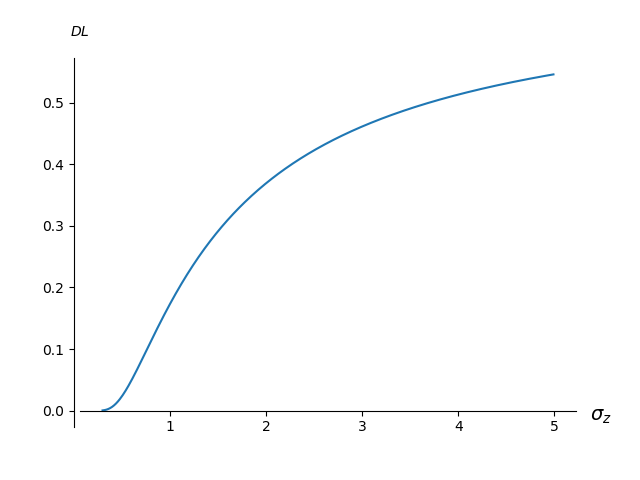}
     \label{fig:meanpos}}
     \subfloat[][$\mu_f-\mu_c=-1$]{\includegraphics[width=0.25\textwidth]{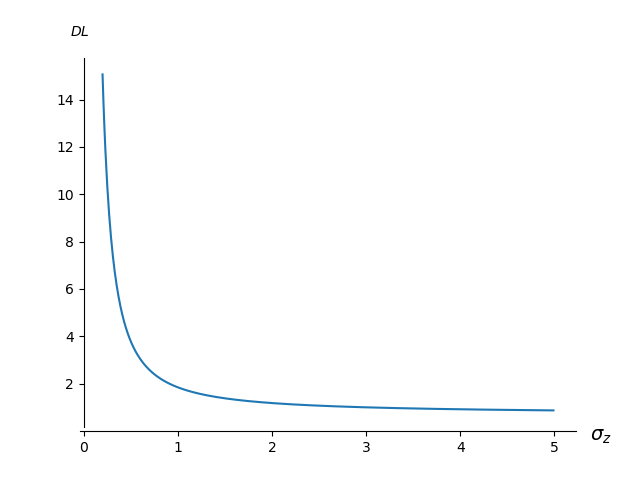}
     \label{fig:meanneg}}
     \caption{DL Loss vs $\sqrt{\sigma_f^2+\sigma_c^2}$. (a) $\mu_f-\mu_c$ is fixed to 1, i.e. the model estimates $\mu_f$ correctly as greater than $\mu_c$; (b) $\mu_f-\mu_c$ is fixed to -1, i.e. the model estimates $\mu_f$ incorrectly as less than $\mu_c$.}
     \label{fig:fix-mean-var-sigma}
\end{figure}

To see the effect of model's uncertainty predictions on the DL loss, we fix the $\mu_f-\mu_c$ to $1$ and $-1$, where the model's estimation is correct and incorrect, respectively. Figure~\ref{fig:meanpos} shows that when the model's estimation is correct, increasing the uncertainty increases the loss. Similarly,  Figure~\ref{fig:meanneg} shows that the DL loss decreases as the uncertainty increases when the model's estimation is incorrect. To sum up, our loss encourages confidence when the estimation is correct, encourages uncertainty when the estimation is incorrect and to decrease the loss, the model would learn to predict the correct confidence in its estimations.

\textbf{Derivatives of the DL loss function} First, we examine the derivatives of our loss function in Equation~(\ref{eq:dist_loss}). Its derivatives with respect to $\mu_f$ and $\mu_c$ are given in Equation~(\ref{eq:derivative_mu}).  Note that they only differ in sign, means that they always change in opposite direction. Therefore, we only plot and discuss gradients of $\mu_f$.

\begin{equation}
\label{eq:derivative_mu}
    \begin{aligned}
    \frac{\partial ~DL}{\partial \mu_f} &=
    -\frac{
    \sqrt{\frac{2}{\pi}} e^{ \frac{-\mu_z^2}{2\sigma_z^2} }
    }
    {
    \sigma_z \left( 1 - \Phi\left( \frac{-\mu_z}{\sqrt{2} \sigma_z} \right) \right)
    } 
    \\
    \frac{\partial ~DL}{\partial \mu_c} &=
    \frac{
    \sqrt{\frac{2}{\pi}} e^{ \frac{-\mu_z^2}{2\sigma_z^2} }
    }
    {
    \sigma_z \left( 1 - \Phi\left( \frac{-\mu_z}{\sqrt{2} \sigma_z} \right) \right)
    }
    \end{aligned}
\end{equation}

\begin{figure}
  \centering
     \subfloat[][$\sigma_z = 1$]{\includegraphics[width=0.4\textwidth]{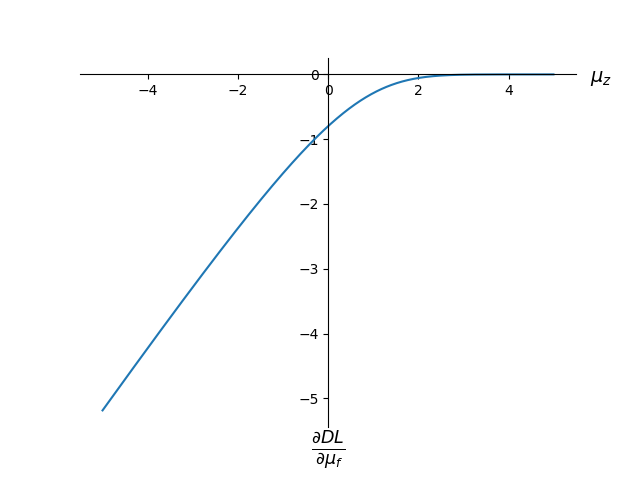}
     \label{fig:fix-sig-grad-muf-by-muz}}
     \\
     \subfloat[][$\mu_z=1$]{\includegraphics[width=0.23\textwidth]{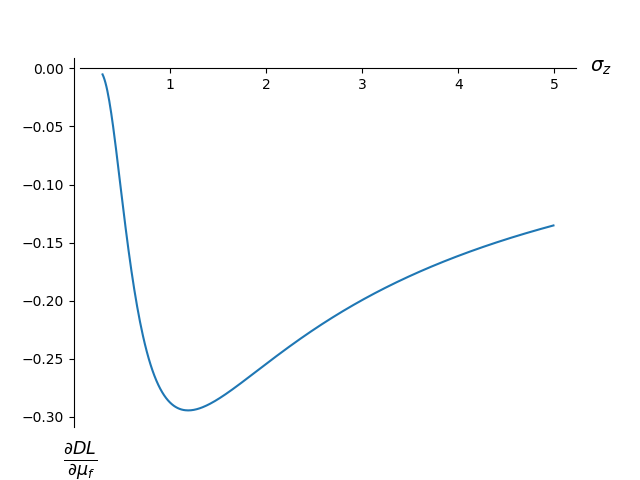}
     \label{fig:fix-muz1-grad-muf-by-sig}}
     \subfloat[][$\mu_z=-1$]{\includegraphics[width=0.23\textwidth]{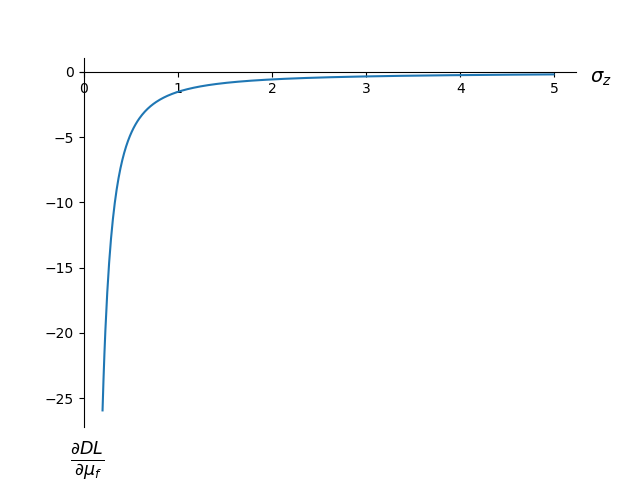}
     \label{fig:fix-muz-1-grad-muf-by-sig}}
     \caption{ Gradients of $\mu_f$ vs. $\mu_z$ (a), $\sigma_z$ (b,c) .}
     \label{fig:grad-muf}
\end{figure}

Figure~\ref{fig:grad-muf} shows the gradients of $\mu_f$ by fixing the $\sigma_z$ and plotting it against $\mu_z$ (Fig.~\ref{fig:fix-sig-grad-muf-by-muz}), and by fixing the $\mu_z$ to $+1$ and $-1$ and plotting it against $\sigma_z$ (Fig.~\ref{fig:fix-muz1-grad-muf-by-sig} and \ref{fig:fix-muz-1-grad-muf-by-sig}, respectively). Since $\mu_f$ is the estimated mean for the pixel that is supposed to be farther away, it receives only negative gradients which increases its value. Figure~\ref{fig:fix-sig-grad-muf-by-muz} shows that the absolute value of the gradients increases as the mistake, $|\mu_z| \text{ when } \mu_z < 0$,  gets bigger, and it vanishes when the estimation is corrected, $\mu_z > 0$, reflecting the behaviour of the loss function shown in Figure~\ref{fig:fix-sigma-var-mean}. When we fix the $\mu_z$ to $1$ where the model's prediction is correct (Fig.~\ref{fig:fix-muz1-grad-muf-by-sig}), the absolute value of gradients which $\mu_f$ receives increases with the uncertainty to some extent meaning that it increases $\mu_f$ more and more aggressively, which is intuitively as expected. Then the rate of the gradient decreases and levels off while the uncertainty keeps increasing. This behaviour is open to interpretation. As the confidence decreases, this loss updates the mean score less aggressively than its value at starting points.
When the model's prediction is wrong (Fig.~\ref{fig:fix-muz-1-grad-muf-by-sig}), $\mu_f$ changes rapidly if the model is confident and the speed of change decreases as the confidence decreases. Again, this reflects the behaviour of the DL loss that can be seen in Figure~\ref{fig:meanneg}.

Equation~(\ref{eq:derivative_sigma}) shows the derivatives with respect to $\sigma_f$ and $\sigma_c$, which are equal except $\frac{\partial DL}{\partial \sigma_f}$ scales with $\sigma_f$ and $\frac{\partial DL}{\partial \sigma_c}$ scales with $\sigma_c$. Note that the sign of the gradients depend on $\mu_z$. When the model's prediction is correct, $\mu_z > 0$, both $\sigma_f$ and $\sigma_c$ receives positive gradients and confidence increases or vice versa. Therefore, we only plot and discuss the gradients of $\sigma_f$.

\begin{equation}
\label{eq:derivative_sigma}
    \begin{aligned}
    \frac{\partial DL}{\partial \sigma_f} &=
    \mu_z \sigma_f \frac{
    \sqrt{\frac{2}{\pi}} e^{-\frac{\mu_z^2}{2\sigma_z^2}}
    }
    { \sigma_z^{3} \left(1-\Phi\left(\frac{-\mu_z}{\sqrt{2} \sigma_z}\right)\right)}
    \\
    \frac{\partial DL}{\partial \sigma_c} &=
    \mu_z \sigma_c \frac{
    \sqrt{\frac{2}{\pi}} e^{-\frac{\mu_z^2}{2\sigma_z^2}}
    }
    { \sigma_z^{3} \left(1-\Phi\left(\frac{-\mu_z}{\sqrt{2} \sigma_z}\right)\right)}
    \end{aligned}
\end{equation}

\begin{figure}
     \centering
     \subfloat[][$\sigma_f = \sigma_c = 1$]{\includegraphics[width=0.4\textwidth]{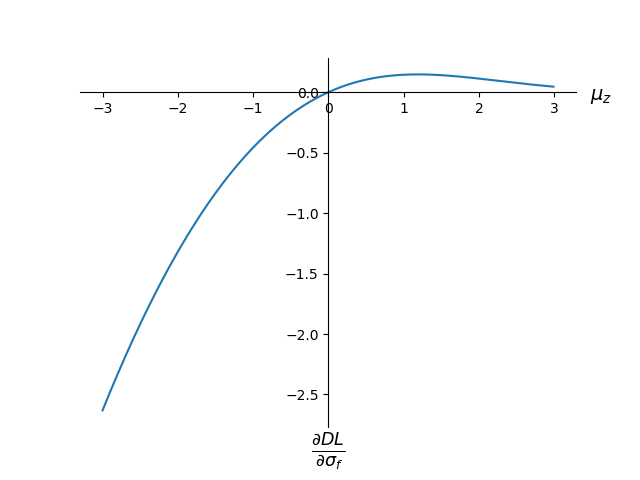}
     \label{fig:grad-sigmaf-by-muz}}
     \\
     \subfloat[][$\mu_z=1$, $\sigma_f = 1$]{\includegraphics[width=0.23\textwidth]{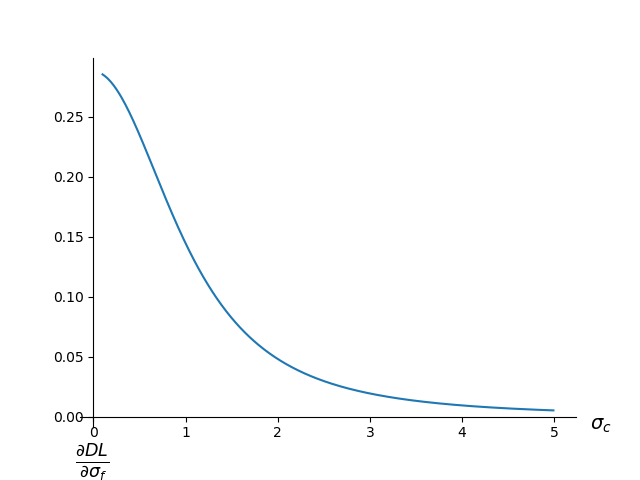}
     \label{fig:grad-sigmaf-by-sigmac-posmuz}}
     \subfloat[][$\mu_z=-1$, $\sigma_f = 1$]{\includegraphics[width=0.23\textwidth]{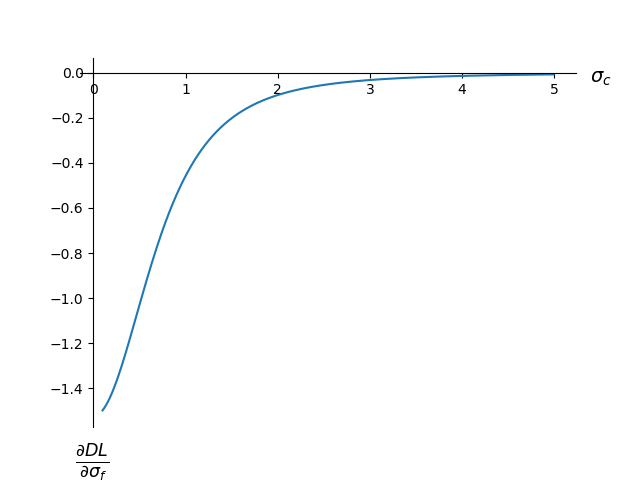}
     \label{fig:grad-sigmaf-by-sigmac-negmuz}}
     \caption{ Gradients of $\sigma_f$ vs. $\mu_z$ (a), $\sigma_c$ (b,c) .}
     \label{fig:grad-sigmaf}
\end{figure}

Figure~\ref{fig:grad-sigmaf} shows the gradients of $\sigma_f$ by fixing the $\sigma_z$ and plotting it against $\mu_z$ (Fig.~\ref{fig:grad-sigmaf-by-muz}), and by fixing the $\mu_z$ to $+1$ and $-1$ and plotting it against $\sigma_c$ (Fig.~\ref{fig:grad-sigmaf-by-sigmac-posmuz} and \ref{fig:grad-sigmaf-by-sigmac-negmuz}, respectively). As it can be seen in Figure~\ref{fig:grad-sigmaf-by-muz}, when the model's prediction is incorrect, $\mu_z < 0$, increase in the uncertainty increases as the mistake, $|\mu_z|$, gets bigger. However, when the model's prediction is correct, $\mu_z > 0$, the decrease in the uncertainty vanishes as $\mu_z$ increases since as long as $P(d_f > d_c) > 0.5$, increase in the $\mu_z$ is only rewarded to some extend by the loss function (see Fig.~\ref{fig:fix-sigma-var-mean}). Furthermore, in both plots of (b) and (c), $\frac{\partial DL}{\partial\sigma_f}$ vanishes as $\sigma_c$ increases, implying that the information gain decreases as the uncertainty increases.

\section{Experiments and Results}
We divide our experiments into three categories. 
First, we conduct an analysis where we examine the confidence prediction on the DIW test split, and devise an experiment to empirically show the usefulness of confidence prediction for the metric depth estimation task. Next, we compare learning standard deviation $\sigma$ directly and reciprocal of it $\frac{1}{\sigma}$. Lastly, we compare our proposed approach with state of the art baselines to show our method's performance for the relative depth estimation task.

For relative depth estimation tasks, we report weighted human disagreement rate (WHDR) \cite{zoran_learning_2015} as
\begin{itemize}
    \item $ \text{WHDR} =  \frac{\sum_{i j} \omega_{i j} \mathbf{1}\left(r_{i j} \neq \tilde{r}_{i j}\right)}{\sum_{i j} \omega_{i j}}$,
\end{itemize}
where $\omega_{i j}$ is the human confidence weight and set to $1$ for DIW test split, $r_{i j}$ and $\tilde{r}_{i j}$ represent ground truth and predicted ordinal relations, respectively. For metric depth estimation tasks, we report the metrics from \cite{eigen_depth_2014}.


We measure calibration performance, which is the degree of consistency between model's predicted probabilities of outcomes and the true probabilities of those outcomes.
To this end, we use expected calibration error (ECE) \cite{naeini_obtaining_2015}, its variants maximum calibration error (MCE) \cite{naeini_obtaining_2015} and adaptive ECE (AdaECE) \cite{mukhoti_calibrating_2020}, and reliability plots \cite{niculescu-mizil_predicting_2005}. They are defined for classification settings where the model's probability output (interpreted as confidence) for a held out data set with $N$ instances is investigated. To calculate ECE, the probability interval $[0,1]$ is divided into $M$ bins and test instances are divided into each bin based on the model's confidence. Let $A_i$ be the average accuracy at bin $i$, $B_i$ be the number of items at bin $i$, and $C_i$ be the average confidence at bin $i$. The aforementioned measures are calculated as follows:
\begin{itemize}
    \item ECE = $ \sum_{i=1}^{M} \frac{\left|B_{i}\right|}{N}\left|A_{i}-C_{i}\right| $
    \item MCE = $\max _{i \in\{1, \ldots, M\}}\left|A_{i}-C_{i}\right|$
    \item AdaECE = $\sum_{i=1}^{M} \frac{\left|B_{i}\right|}{N}\left|A_{i}-C_{i}\right| \text { s.t. } \forall i, j \cdot\left|B_{i}\right|=\left|B_{j}\right|$
    \item reliability plots: plot of accuracies at each bin as a bar chart.
\end{itemize}

All the experiments are conducted with the same setting. We perform on-the-fly data augmentation. Specifically, we horizontally flip the image, rescale and crop it to the input size of 384$\times$384, and apply rotation. We experiment with a common architecture used in previous works: EncDecResNet \cite{xian_monocular_2018}. We use stochastic gradient descent (SGD) with cosine annealing learning rate scheduler \cite{loshchilov_sgdr_2016} which cyclically changes learning rate between $[1e-3, 1e-7]$ and completes a cycle at every 5 epoch. Batch size is 8 in all experiments. All of the hyperparameters are chosen heuristically due to limited computational resources. 

\subsection{Investigating confidence prediction}

To demonstrate the effectiveness of our approach for confidence prediction, we measure the model calibration of EncDecResNet trained with DL loss on RedWeb training set, which we refer to as \textit{DL\_EDR}. We choose $r_{ij}=1$ as positive class and calculate the class probability $P(r_{ij}=1)$ as $P(d_i>d_j)$, as given in Equation~(\ref{eq:prob_Q}).  
However, this probability is affected by the $\mu_i$ and $\mu_j$ predictions as well. To better show the effect of confidence prediction, we also use $\mu_i - \mu_j$ as $P(r_{ij}=1)$ by mapping it to $[0,1]$ range (represented as $P(r_{ij}=1)=\mu_i-\mu_j$). The latter formulation essentially calculates the confidence based on the distances between the estimated centers of both pixels' depth distribution, \ie confidence increases as the distances between the centers increases.

\begin{table}[]
\centering
\caption{Calibration measures of \textit{DL\_EDR} on DIW test split. First row shows the measures calculated with $\mu$ predictions only. Second row shows the measures calculated with $\sigma$ predictions as well as $\mu$ predictions. Lower is better. }
\begin{tabular}{lccc}
\hline
$P(r_{ij}=1)=$                  & ECE       & AdaECE    & MCE       \\ \hline
$\mu_i-\mu_j$                   & 0.25      & 0.27      & 0.43 \\
Equation~(\ref{eq:prob_Q})      &\textbf{0.02}&\textbf{0.02}& \textbf{0.05} \\\hline
\end{tabular}
\label{tab:calibration_measures}
\end{table}

\begin{figure}
     \centering
     \includegraphics[width=0.2\textwidth]{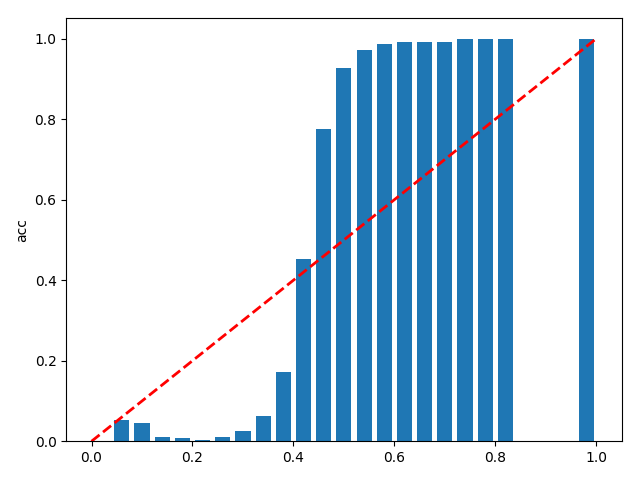}
     \includegraphics[width=0.2\textwidth]{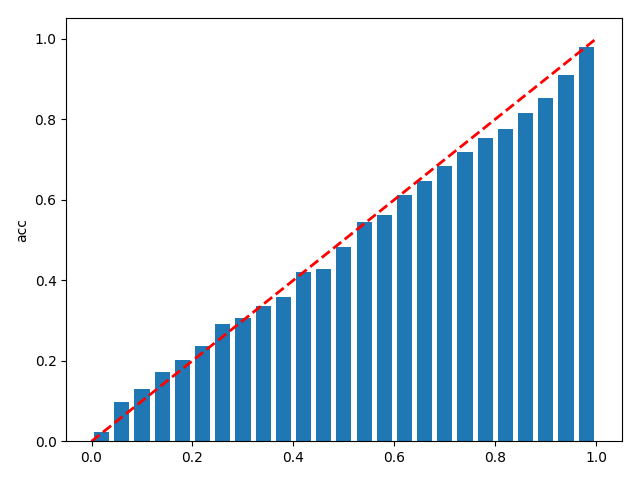}
     \\
     \includegraphics[width=0.2\textwidth]{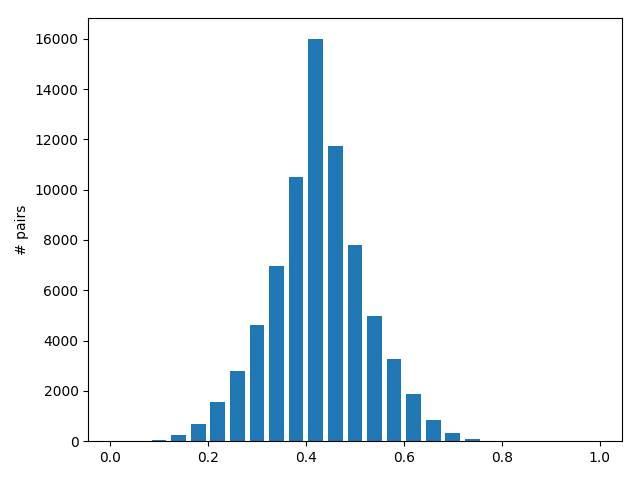}
     \includegraphics[width=0.2\textwidth]{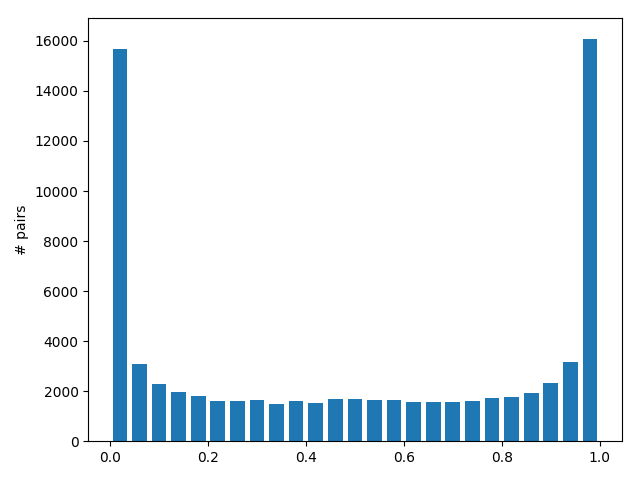}
     \caption{Left column: plots of $P(r_{ij}=1)=\mu_i-\mu_j$. Right column: $P(r_{ij}=1)=$Equation~(\ref{eq:prob_Q}). Top row: reliability plots with 25 bins (following the diagonal line is better). Bottom row: histograms showing how many pairs fall into each confidence bin. }
     \label{fig:reliability_plots}
\end{figure}

Table~\ref{tab:calibration_measures} shows the calibration measures ECE \cite{naeini_obtaining_2015}, AdaECE \cite{mukhoti_calibrating_2020}, and MCE \cite{naeini_obtaining_2015}, and Figure~\ref{fig:reliability_plots} shows the reliability plots \cite{niculescu-mizil_predicting_2005} of \textit{DL\_EDR} with different interpretations of $P(r_{ij}=1)$. Both calibration measures and reliability plots indicate that $\mu_i - \mu_j$ alone, is not a good indicator of confidence. However, when we utilize confidence predictions as well, model output becomes almost perfectly calibrated, indicating that confidence predictions reflects the model's expected accuracy.

To empirically show the usefulness of confidence information, we conduct the following experiment. We train the EncDecResNet of \cite{xian_monocular_2018} on the NYUDv2 dataset for the absolute depth estimation task. First, we train the network using RGB images and the corresponding score map (SM) estimation from \textit{DL\_EDR} as an input (RGB+SM, abbreviated as \textit{W/O CM}). Next, we also input the confidence map (CM) prediction of \textit{DL\_EDR} (RGB+SM+CM, abbreviated as \textit{W/ CM}) and repeat the training. Table~\ref{tab:nyu_conf} shows the results. Inputting confidence map alongside other inputs increases the performance in most of the metrics, especially the most challenging accuracy, $thr=1.25$. We conjecture that confidence predictions allow network to employ different strategies for parts that are likely to have wrong relative depth score, hence the performance gain.

\begin{table}[]
\centering
\caption{Results of experiments conducted on NYUDv2 with and without confidence map as an input. }
\small\addtolength{\tabcolsep}{-3pt}
\scalebox{0.8}{%
\begin{tabular}{lccccccc}
\hline
          & \multicolumn{3}{c}{Accuracy}                                                    & \multicolumn{4}{c}{Error}                                     \\ \hline
          & $thr = 1.25$      & $thr = 1.25^2$ & $thr = 1.25^3$ & \begin{tabular}[c]{@{}c@{}}RMSE \\ (linear)\end{tabular} & \begin{tabular}[c]{@{}c@{}}RMSE \\ (log)\end{tabular}    & absrel        & sqrrel        \\ \hline
W/O CM    & 70.4\%          & 92.8\%                        & \textbf{98.4\%}                        & 0.73          & 0.26          & \textbf{0.19}          & 0.16          \\
W/ CM & \textbf{71.78\%} & \textbf{92.9\%}               & 98.3\%               & \textbf{0.70} & \textbf{0.25} & \textbf{0.19} & \textbf{0.15} \\ \hline
\end{tabular}}
\label{tab:nyu_conf}
\end{table}

\subsection{Learning confidence vs. uncertainty}
We also conduct experiments to empirically show the effectiveness of learning reciprocal of the standard deviation, $\frac{1}{\sigma}$. We train EncDecResNet on RedWeb dataset and experiment with learning standard deviation. In this case, we can interpret the second channel of the output as uncertainty map. We also repeat the same experiment with treating networks second output as $\frac{1}{\sigma}$ in loss formulation. In this case, we can interpret the second channel of the output as confidence map. Table~\ref{tab:confvsunc} shows the results which indicate that learning confidence performs much better when compared to learning uncertainty. In the uncertainty version of the model, we observe that standard deviation output of the model diverges very quickly and the model converges to a local optima.

\begin{table}[]
\centering
\caption{WHDR on DIW test split with different interpretation of the second output of the network.}
\begin{tabular}{lll}
\hline
Interpretation     & Uncertainty & Confidence \\ \hline
WHDR &    31.63\%         & \textbf{16.15\%}    \\ \hline
\end{tabular}
\label{tab:confvsunc}
\end{table}

\subsection{Comparison with state of the art}
\captionsetup[subfigure]{labelformat=empty}
\begin{figure*}
\begin{center}
\subfloat{ \includegraphics[width=0.085\linewidth]{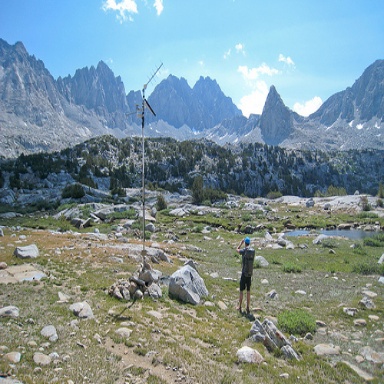} }\hspace{-0.2cm}
\subfloat{ \includegraphics[width=0.085\linewidth]{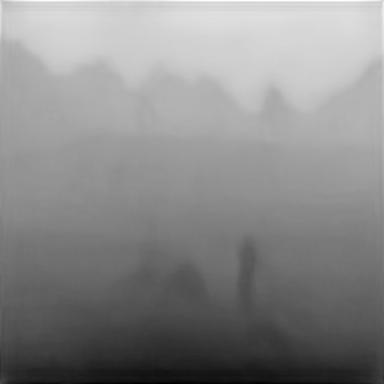} }\hspace{-0.2cm}
\subfloat{ \includegraphics[width=0.085\linewidth]{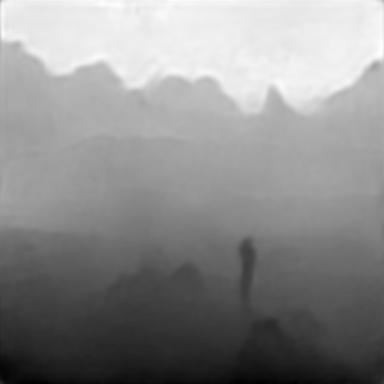} }\hspace{-0.2cm}
\subfloat{ \includegraphics[width=0.085\linewidth]{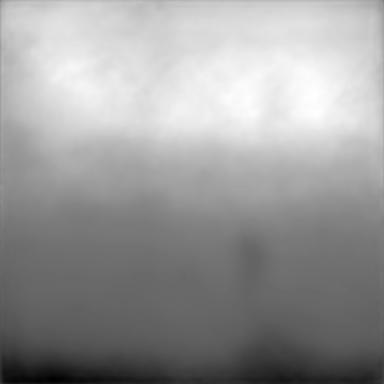} }\hspace{-0.2cm}
\subfloat{ \includegraphics[width=0.085\linewidth]{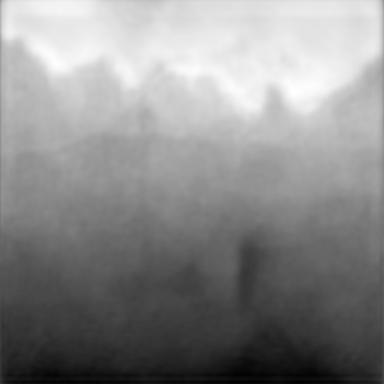} }\hspace{-0.2cm}
\subfloat{ \includegraphics[width=0.085\linewidth]{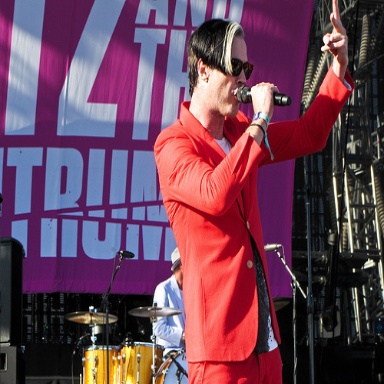} }\hspace{-0.2cm}
\subfloat{ \includegraphics[width=0.085\linewidth]{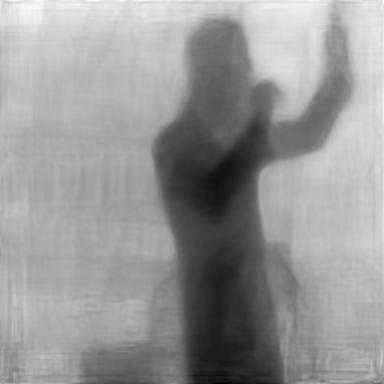} }\hspace{-0.2cm}
\subfloat{ \includegraphics[width=0.085\linewidth]{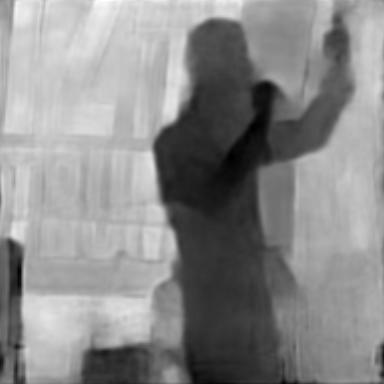} }\hspace{-0.2cm}
\subfloat{ \includegraphics[width=0.085\linewidth]{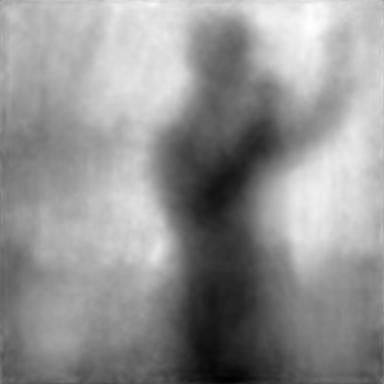} }\hspace{-0.2cm}
\subfloat{ \includegraphics[width=0.085\linewidth]{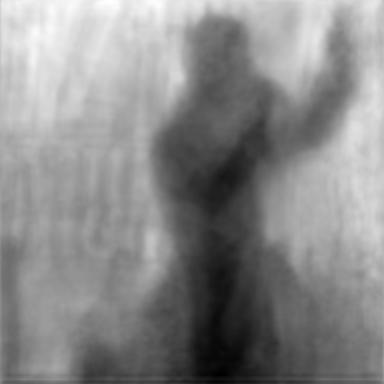} }\hspace{-0.2cm}
\\[-2ex]
\subfloat{ \includegraphics[width=0.085\linewidth]{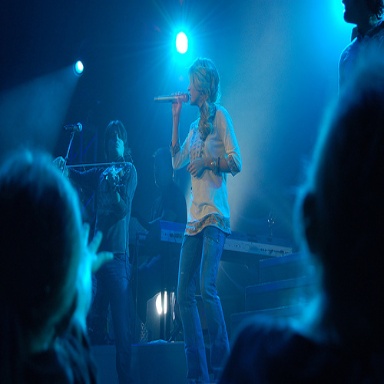} }\hspace{-0.2cm}
\subfloat{ \includegraphics[width=0.085\linewidth]{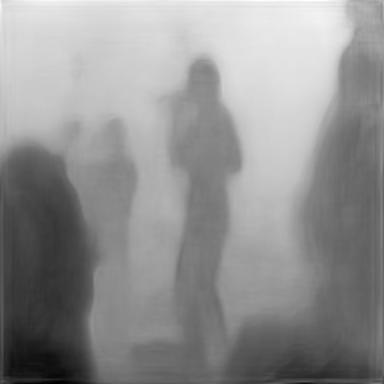} }\hspace{-0.2cm}
\subfloat{ \includegraphics[width=0.085\linewidth]{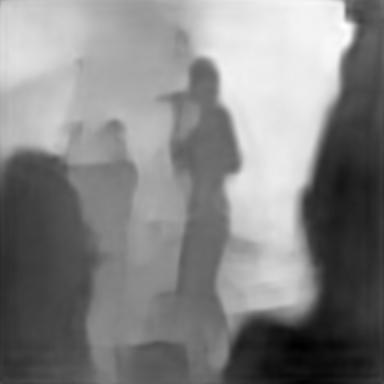} }\hspace{-0.2cm}
\subfloat{ \includegraphics[width=0.085\linewidth]{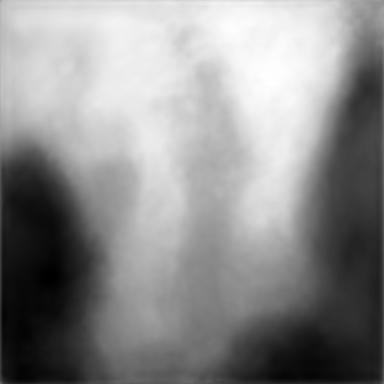} }\hspace{-0.2cm}
\subfloat{ \includegraphics[width=0.085\linewidth]{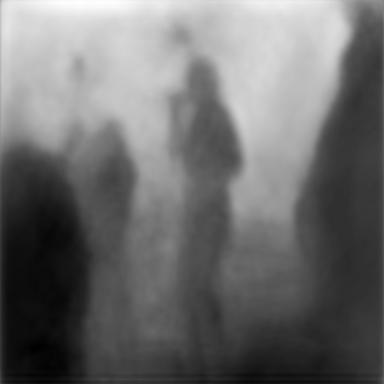} }\hspace{-0.2cm}
\subfloat{ \includegraphics[width=0.085\linewidth]{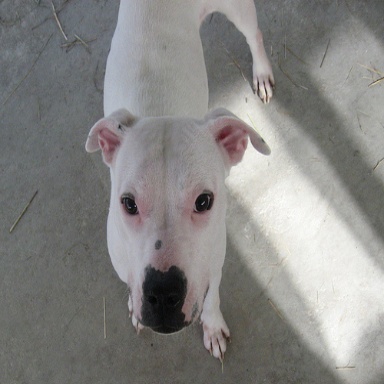} }\hspace{-0.2cm}
\subfloat{ \includegraphics[width=0.085\linewidth]{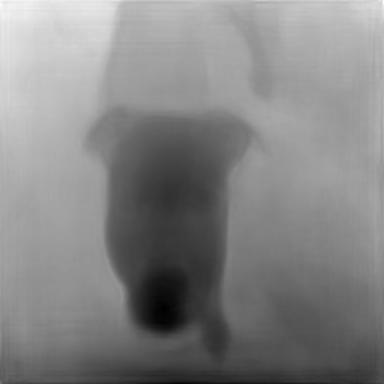} }\hspace{-0.2cm}
\subfloat{ \includegraphics[width=0.085\linewidth]{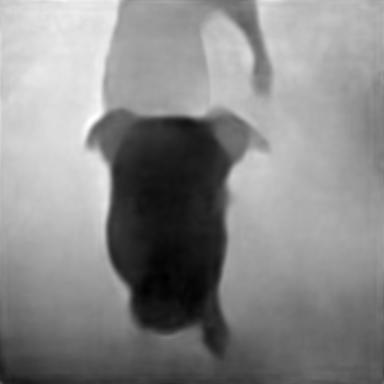} }\hspace{-0.2cm}
\subfloat{ \includegraphics[width=0.085\linewidth]{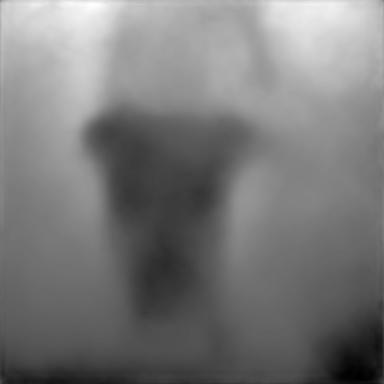} }\hspace{-0.2cm}
\subfloat{ \includegraphics[width=0.085\linewidth]{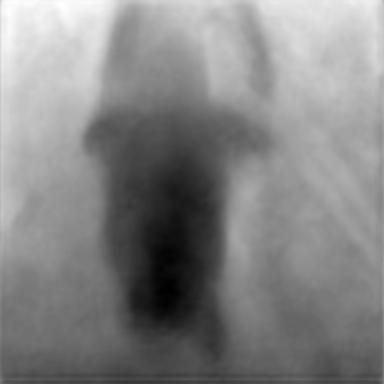} }\hspace{-0.2cm}
\\[-2ex]
\subfloat{ \includegraphics[width=0.085\linewidth]{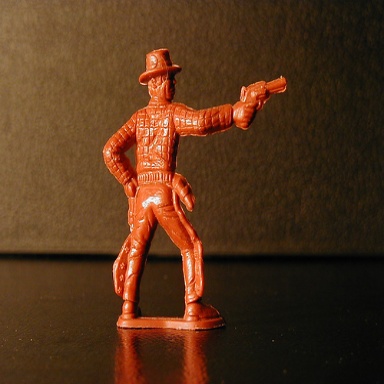} }\hspace{-0.2cm}
\subfloat{ \includegraphics[width=0.085\linewidth]{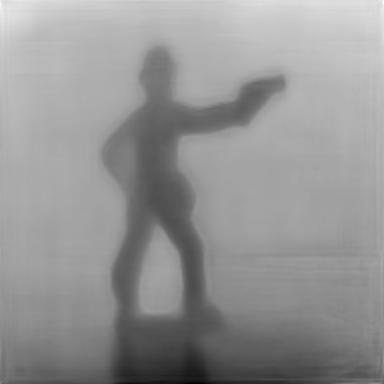} }\hspace{-0.2cm}
\subfloat{ \includegraphics[width=0.085\linewidth]{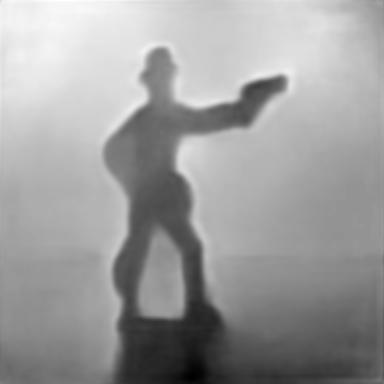} }\hspace{-0.2cm}
\subfloat{ \includegraphics[width=0.085\linewidth]{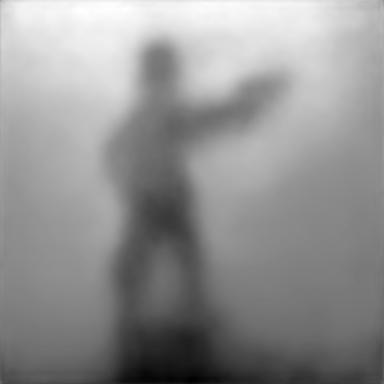} }\hspace{-0.2cm}
\subfloat{ \includegraphics[width=0.085\linewidth]{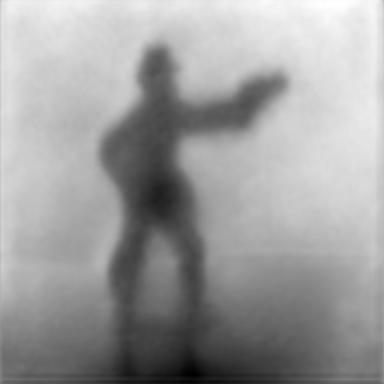} }\hspace{-0.2cm}
\subfloat{ \includegraphics[width=0.085\linewidth]{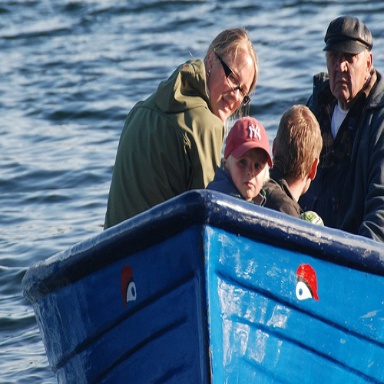} }\hspace{-0.2cm}
\subfloat{ \includegraphics[width=0.085\linewidth]{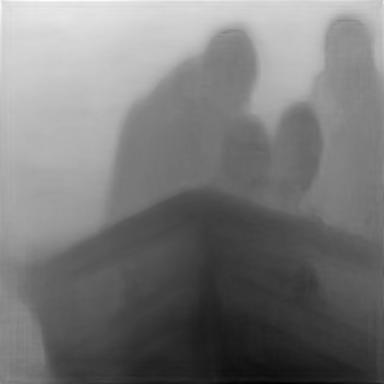} }\hspace{-0.2cm}
\subfloat{ \includegraphics[width=0.085\linewidth]{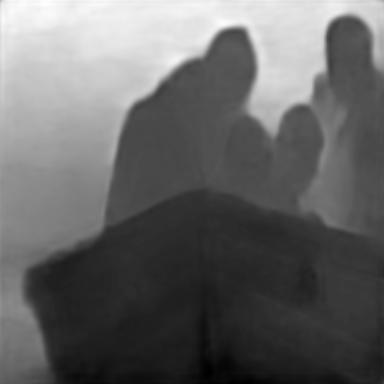} }\hspace{-0.2cm}
\subfloat{ \includegraphics[width=0.085\linewidth]{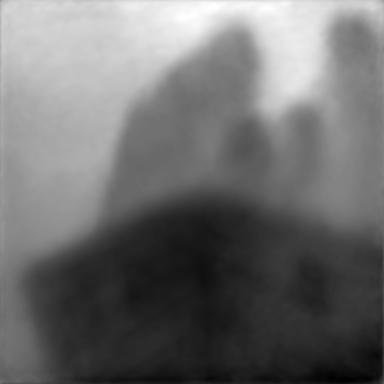} }\hspace{-0.2cm}
\subfloat{ \includegraphics[width=0.085\linewidth]{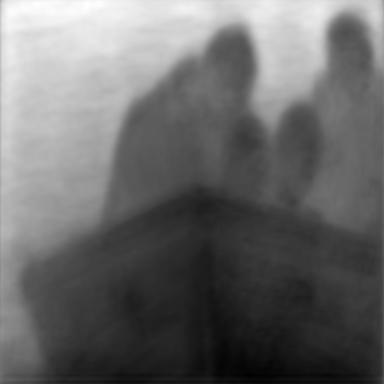} }\hspace{-0.2cm}
\\[-2ex]
\subfloat{ \includegraphics[width=0.085\linewidth]{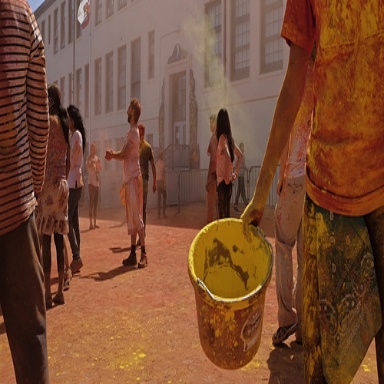} }\hspace{-0.2cm} \vspace{-0.4cm}
\subfloat[RedWeb +DIW]{ \includegraphics[width=0.085\linewidth]{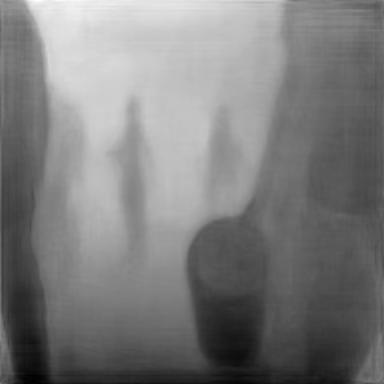} }\hspace{-0.2cm}
\subfloat[RedWeb]{ \includegraphics[width=0.085\linewidth]{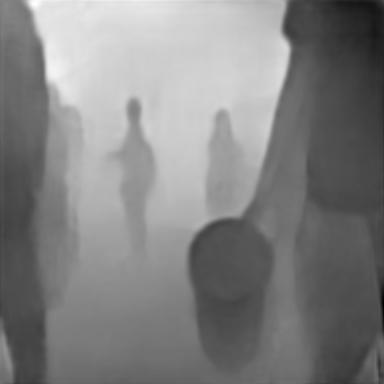} }\hspace{-0.2cm}
\subfloat[YouTube3D]{ \includegraphics[width=0.085\linewidth]{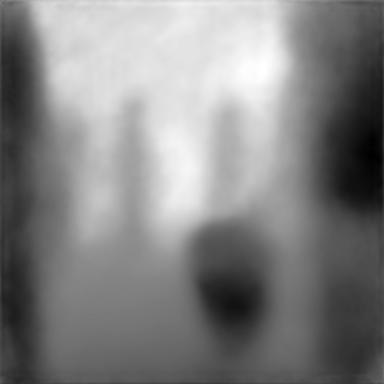} }\hspace{-0.2cm}
\subfloat[DIW]{ \includegraphics[width=0.085\linewidth]{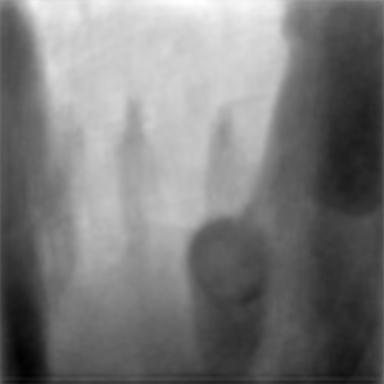} }\hspace{-0.2cm}
\subfloat{ \includegraphics[width=0.085\linewidth]{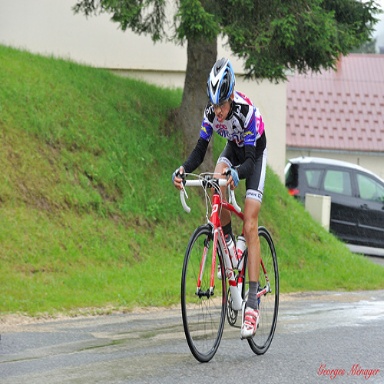} }\hspace{-0.2cm}
\subfloat[RedWeb +DIW]{ \includegraphics[width=0.085\linewidth]{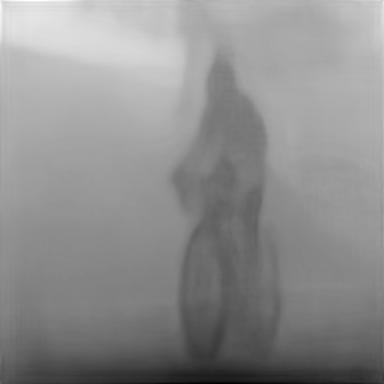} }\hspace{-0.2cm}
\subfloat[RedWeb]{ \includegraphics[width=0.085\linewidth]{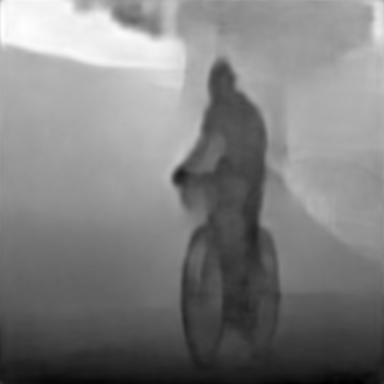} }\hspace{-0.2cm}
\subfloat[YouTube3D]{ \includegraphics[width=0.085\linewidth]{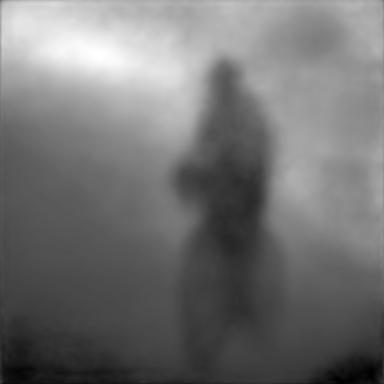} }\hspace{-0.2cm}
\subfloat[DIW]{ \includegraphics[width=0.085\linewidth]{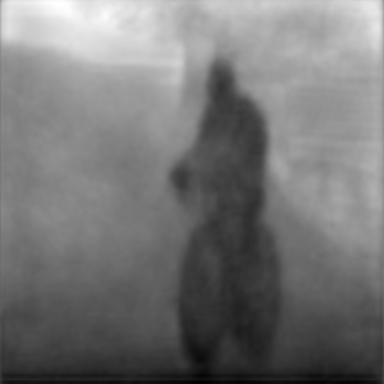} }\hspace{-0.2cm}
\end{center}
\caption{Qualitative results of \textit{DL\_EDR} trained on different training sets.}
\label{fig:qualitiative}
\end{figure*}

\begin{figure}
\begin{center}
\subfloat{ \includegraphics[width=0.25\linewidth]{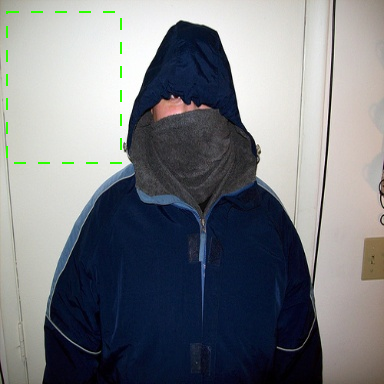} }\hspace{-0.2cm}
\subfloat{ \includegraphics[width=0.25\linewidth]{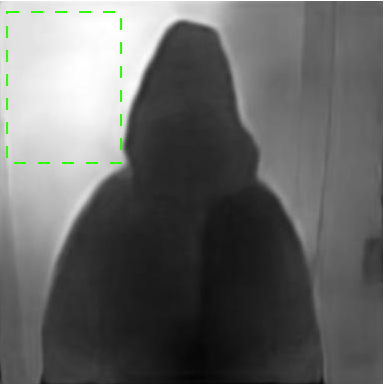} }\hspace{-0.2cm}
\subfloat{ \includegraphics[width=0.25\linewidth]{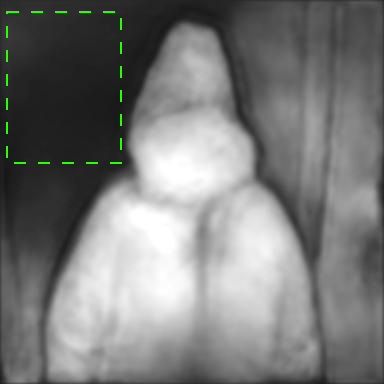} }\hspace{-0.2cm}
\\[-2ex]
\subfloat{ \includegraphics[width=0.25\linewidth]{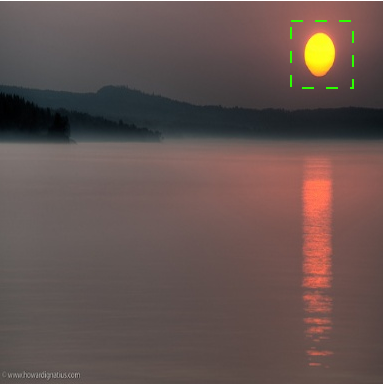} }\hspace{-0.2cm}
\subfloat{ \includegraphics[width=0.25\linewidth]{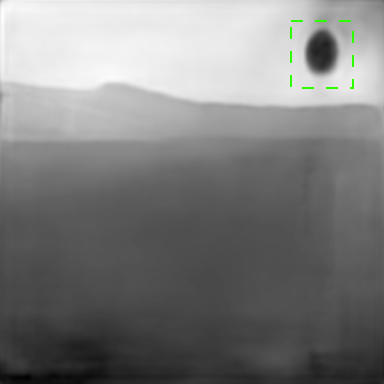} }\hspace{-0.2cm}
\subfloat{ \includegraphics[width=0.25\linewidth]{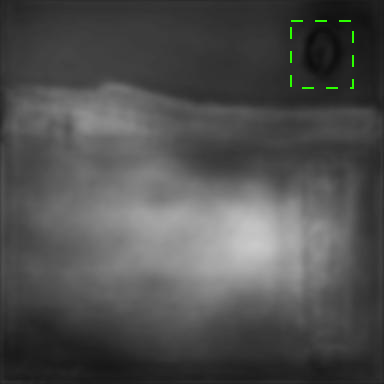} }\hspace{-0.2cm}
\\[-2ex]
\subfloat{ \includegraphics[width=0.25\linewidth]{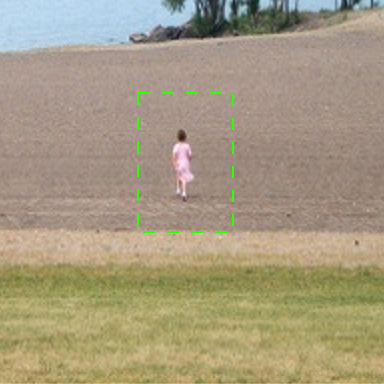} }\hspace{-0.2cm}
\subfloat{ \includegraphics[width=0.25\linewidth]{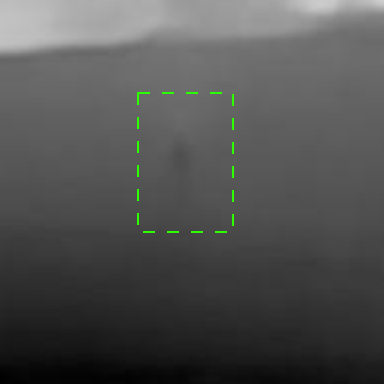} }\hspace{-0.2cm}
\subfloat{ \includegraphics[width=0.25\linewidth]{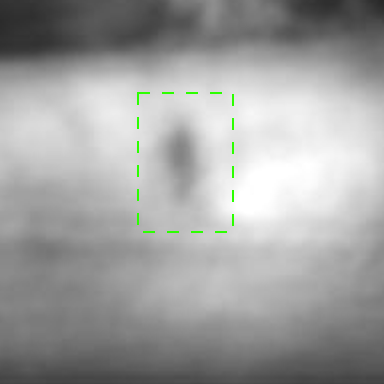} }\hspace{-0.2cm}
\\[-2ex] 
\subfloat[RGB]{ \includegraphics[width=0.25\linewidth]{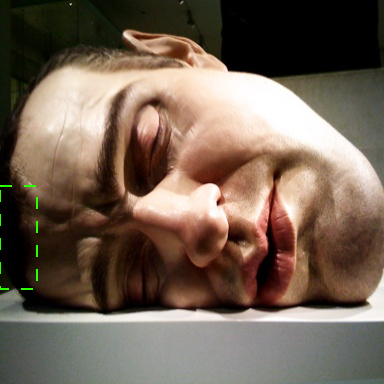} }\hspace{-0.2cm} \vspace{-0.4cm}
\subfloat[Prediction]{ \includegraphics[width=0.25\linewidth]{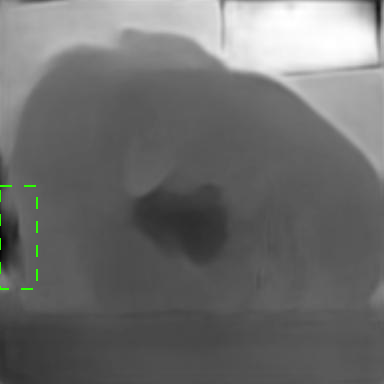} }\hspace{-0.2cm}
\subfloat[Confidence]{ \includegraphics[width=0.25\linewidth]{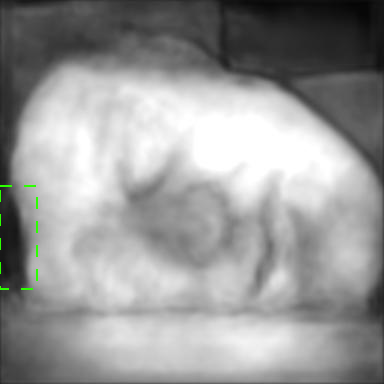} }\hspace{-0.2cm}
\end{center}
\caption{Qualitative results of confidence predictions of \textit{DL\_EDR} on DIW test split. Green boxes indicate important parts of the images. Better viewed digitally.}
\label{fig:confidence_qualitative}
\end{figure}
To show the effectiveness of our approach for the relative depth estimation task, we compare our results with several similar works \cite{chen_single-image_2016, xian_monocular_2018, chen_learning_2019}. We use the following naming convention in method names for ease of understanding. The first part indicates the original work that publishes the given result. The second part represents the neural network model, \textit{HG} being Hourglass \cite{chen_single-image_2016} and \textit{EDR} being EncDecResNet \cite{xian_monocular_2018}. To train \textit{EDR} model with the proposed approach, we only add an additional head with Leaky ReLU activation for confidence output without any further modification. The last part represents the loss that is used where \textit{R} is the ranking loss in \cite{chen_single-image_2016}, \textit{IR} is the improved ranking loss in \cite{xian_monocular_2018}, and \textit{DL} is the proposed distributional loss. Following previous works \cite{xian_monocular_2018, chen_learning_2019}, we use ImageNet pretrained weights for initializing encoder part of the \textit{EDR} in all experiments.

\begin{table}[]
\centering
\caption{Comparisons with various baselines from the literature. Results are from DIW test split. 
}
\small\addtolength{\tabcolsep}{-3pt}
\begin{tabular}{clc}
\hline
Training set               & \multicolumn{1}{c}{Method} & WHDR      \\ \hline
\multirow{3}{*}{YouTube3D} & Chen\_HG\_IR \cite{chen_learning_2019}               & 19.01\%   \\
                           & Chen\_EDR\_IR \cite{chen_learning_2019}             & 16.21\%   \\
                           & Ours\_EDR\_DL              & \textbf{16.08\%}          \\ \hline
\multirow{3}{*}{DIW}       & Chen\_HG\_R \cite{chen_single-image_2016}            & 22.14\%   \\
                           & Xian\_EDR\_R \cite{xian_monocular_2018}             & 14.98\%   \\
                           & Ours\_EDR\_DL              & \textbf{12.59\%} \\ \hline
\multirow{2}{*}{RedWeb}    & Chen\_EDR\_IR \cite{chen_learning_2019}             & 16.31\%   \\
                           & Ours\_EDR\_DL              & \textbf{16.15\%}   \\ \hline
\multirow{3}{*}{RedWeb+DIW}    & Xian\_EDR\_IR \cite{xian_monocular_2018}             & 11.37\%   \\
                            & Chen\_EDR\_IR \cite{chen_learning_2019}   & 12.03\%\\
                           & Ours\_EDR\_DL              & \textbf{11.01\%}   \\ \hline\hline
RedWeb+DIW+YouTube3D    & Chen\_EDR\_IR \cite{chen_learning_2019}   & \textbf{10.59\%} \\ \hline
\end{tabular}
\label{tab:baseline}
\end{table}

Table~\ref{tab:baseline} shows results on DIW test split. We achieve state of the art performances on all experimented datasets. We believe our good performances in all datasets with heuristically chosen parameters indicate that our method is not sensitive to hyperparameters. Figure~\ref{fig:qualitiative} shows qualitative examples from DIW test split. We observe that model trained on RedWeb dataset produce sharper results which indicates that number of annotated pixel pairs is important for visually better results since RedWeb dataset has dense annotation. In Figure~\ref{fig:confidence_qualitative}, we see that model has indeed low confidence for parts where the depth estimation is not accurate or for parts where the depth estimation is hard to make such as background or edges. For instance, in the first row of images, although the wall background seems parallel to the camera plane, the estimated depth is larger on the right side of the person than that of the left. This is marked by a lower confidence estimate. Similarly, see the indicated regions (by dashed green boxes) where the confidence map correctly reproduces the expected lower confidence scores.

\section{Conclusion}
In this paper, we propose a new pairwise ranking approach, and illustrate it in the problem of relative depth estimation. We estimate the probability distribution over depth values, and maximize the likelihood of the observed pairwise ordinal relations by training a neural network model with the proposed DL loss. With this formulation, we achieve better or comparable performances with prior art while outputting confidence for estimations as well. The new ranking distributional loss with uncertainty that is presented in this work is not specific to relative depth estimation, and can be utilized in other problems involving ordinal relations between measurements.



{\small
\bibliographystyle{ieee}
\bibliography{depthEstimation}

\begin{thebibliography}{10}\itemsep=-1pt

\bibitem{burges_learning_2005}
C.~Burges, T.~Shaked, E.~Renshaw, A.~Lazier, M.~Deeds, N.~Hamilton, and G.~N.
  Hullender.
\newblock Learning to rank using gradient descent.
\newblock In {\em Proceedings of the 22nd {International} {Conference} on
  {Machine} learning ({ICML}-05)}, pages 89--96, 2005.

\bibitem{cakir_deep_2019}
F.~Cakir, K.~He, X.~Xia, B.~Kulis, and S.~Sclaroff.
\newblock Deep {Metric} {Learning} to {Rank}.
\newblock In {\em Proceedings of the {IEEE} {Conference} on {Computer} {Vision}
  and {Pattern} {Recognition}}, pages 1861--1870, 2019.

\bibitem{cao_estimating_2017}
Y.~Cao, Z.~Wu, and C.~Shen.
\newblock Estimating depth from monocular images as classification using deep
  fully convolutional residual networks.
\newblock {\em IEEE Transactions on Circuits and Systems for Video Technology},
  28(11):3174--3182, 2017.

\bibitem{cao_learning_2007}
Z.~Cao, T.~Qin, T.-Y. Liu, M.-F. Tsai, and H.~Li.
\newblock Learning to rank: from pairwise approach to listwise approach.
\newblock In {\em Proceedings of the 24th international conference on {Machine}
  learning}, pages 129--136. ACM, 2007.

\bibitem{chen_towards_2019}
P.-Y. Chen, A.~H. Liu, Y.-C. Liu, and Y.-C.~F. Wang.
\newblock Towards {Scene} {Understanding}: {Unsupervised} {Monocular} {Depth}
  {Estimation} {With} {Semantic}-{Aware} {Representation}.
\newblock In {\em Proceedings of the {IEEE} {Conference} on {Computer} {Vision}
  and {Pattern} {Recognition}}, pages 2624--2632, 2019.

\bibitem{chen_single-image_2016}
W.~Chen, Z.~Fu, D.~Yang, and J.~Deng.
\newblock Single-image depth perception in the wild.
\newblock In {\em Advances in {Neural} {Information} {Processing} {Systems}},
  pages 730--738, 2016.

\bibitem{chen_learning_2019}
W.~Chen, S.~Qian, and J.~Deng.
\newblock Learning single-image depth from videos using quality assessment
  networks.
\newblock In {\em Proceedings of the {IEEE} {Conference} on {Computer} {Vision}
  and {Pattern} {Recognition}}, pages 5604--5613, 2019.

\bibitem{eigen_predicting_2015}
D.~Eigen and R.~Fergus.
\newblock Predicting depth, surface normals and semantic labels with a common
  multi-scale convolutional architecture.
\newblock In {\em Proceedings of the {IEEE} international conference on
  computer vision}, pages 2650--2658, 2015.

\bibitem{eigen_depth_2014}
D.~Eigen, C.~Puhrsch, and R.~Fergus.
\newblock Depth map prediction from a single image using a multi-scale deep
  network.
\newblock In {\em Advances in neural information processing systems}, pages
  2366--2374, 2014.

\bibitem{fu_deep_2018}
H.~Fu, M.~Gong, C.~Wang, K.~Batmanghelich, and D.~Tao.
\newblock Deep ordinal regression network for monocular depth estimation.
\newblock In {\em Proceedings of the {IEEE} {Conference} on {Computer} {Vision}
  and {Pattern} {Recognition}}, pages 2002--2011, 2018.

\bibitem{garg_unsupervised_2016}
R.~Garg, V.~K. BG, G.~Carneiro, and I.~Reid.
\newblock Unsupervised cnn for single view depth estimation: {Geometry} to the
  rescue.
\newblock In {\em European {Conference} on {Computer} {Vision}}, pages
  740--756. Springer, 2016.

\bibitem{godard_unsupervised_2017}
C.~Godard, O.~Mac~Aodha, and G.~J. Brostow.
\newblock Unsupervised monocular depth estimation with left-right consistency.
\newblock In {\em Proceedings of the {IEEE} {Conference} on {Computer} {Vision}
  and {Pattern} {Recognition}}, pages 270--279, 2017.

\bibitem{hoiem_automatic_2005}
D.~Hoiem, A.~A. Efros, and M.~Hebert.
\newblock Automatic photo pop-up.
\newblock In {\em {ACM} transactions on graphics ({TOG})}, volume~24, pages
  577--584. ACM, 2005.

\bibitem{laina_deeper_2016}
I.~Laina, C.~Rupprecht, V.~Belagiannis, F.~Tombari, and N.~Navab.
\newblock Deeper depth prediction with fully convolutional residual networks.
\newblock In {\em 2016 {Fourth} international conference on 3D vision (3DV)},
  pages 239--248. IEEE, 2016.

\bibitem{lan_position-aware_2014}
Y.~Lan, Y.~Zhu, J.~Guo, S.~Niu, and X.~Cheng.
\newblock Position-{Aware} {ListMLE}: {A} {Sequential} {Learning} {Process} for
  {Ranking}.
\newblock In {\em {UAI}}, pages 449--458, 2014.

\bibitem{liu_single_2010}
B.~Liu, S.~Gould, and D.~Koller.
\newblock Single image depth estimation from predicted semantic labels.
\newblock In {\em 2010 {IEEE} {Computer} {Society} {Conference} on {Computer}
  {Vision} and {Pattern} {Recognition}}, pages 1253--1260. IEEE, 2010.

\bibitem{loshchilov_sgdr_2016}
I.~Loshchilov and F.~Hutter.
\newblock Sgdr: {Stochastic} gradient descent with warm restarts.
\newblock {\em arXiv preprint arXiv:1608.03983}, 2016.

\bibitem{mertan_listwise}
A.~Mertan, D.~J. Duff, and G.~Unal.
\newblock Relative depth estimation as a ranking problem.
\newblock In {\em 2020 28th Signal Processing and Communications Applications
  Conference (SIU)}, pages 1--4. IEEE, 2020.

\bibitem{michels_high_2005}
J.~Michels, A.~Saxena, and A.~Y. Ng.
\newblock High speed obstacle avoidance using monocular vision and
  reinforcement learning.
\newblock In {\em Proceedings of the 22nd international conference on {Machine}
  learning}, pages 593--600. ACM, 2005.

\bibitem{mukhoti_calibrating_2020}
J.~Mukhoti, V.~Kulharia, A.~Sanyal, S.~Golodetz, P.~H. Torr, and P.~K. Dokania.
\newblock Calibrating {Deep} {Neural} {Networks} using {Focal} {Loss}.
\newblock {\em arXiv preprint arXiv:2002.09437}, 2020.

\bibitem{naeini_obtaining_2015}
M.~P. Naeini, G.~Cooper, and M.~Hauskrecht.
\newblock Obtaining well calibrated probabilities using bayesian binning.
\newblock In {\em Twenty-{Ninth} {AAAI} {Conference} on {Artificial}
  {Intelligence}}, 2015.

\bibitem{niculescu-mizil_predicting_2005}
A.~Niculescu-Mizil and R.~Caruana.
\newblock Predicting good probabilities with supervised learning.
\newblock In {\em Proceedings of the 22nd international conference on {Machine}
  learning}, pages 625--632, 2005.

\bibitem{pizzoli_remode_2014}
M.~Pizzoli, C.~Forster, and D.~Scaramuzza.
\newblock {REMODE}: {Probabilistic}, monocular dense reconstruction in real
  time.
\newblock In {\em 2014 {IEEE} {International} {Conference} on {Robotics} and
  {Automation} ({ICRA})}, pages 2609--2616. IEEE, 2014.

\bibitem{qi_geonet_2018}
X.~Qi, R.~Liao, Z.~Liu, R.~Urtasun, and J.~Jia.
\newblock Geonet: {Geometric} neural network for joint depth and surface normal
  estimation.
\newblock In {\em Proceedings of the {IEEE} {Conference} on {Computer} {Vision}
  and {Pattern} {Recognition}}, pages 283--291, 2018.

\bibitem{qin_query-level_2008}
T.~Qin, X.-D. Zhang, M.-F. Tsai, D.-S. Wang, T.-Y. Liu, and H.~Li.
\newblock Query-level loss functions for information retrieval.
\newblock {\em Information Processing \& Management}, 44(2):838--855, 2008.
\newblock ISBN: 0306-4573 Publisher: Elsevier.

\bibitem{ranjan_competitive_2018}
A.~Ranjan, V.~Jampani, L.~Balles, K.~Kim, D.~Sun, J.~Wulff, and M.~J. Black.
\newblock Competitive {Collaboration}: {Joint} {Unsupervised} {Learning} of
  {Depth}, {Camera} {Motion}, {Optical} {Flow} and {Motion} {Segmentation}.
\newblock {\em arXiv preprint arXiv:1805.09806}, 2018.

\bibitem{ren_cross-domain_2018}
Z.~Ren and Y.~Jae~Lee.
\newblock Cross-domain self-supervised multi-task feature learning using
  synthetic imagery.
\newblock In {\em Proceedings of the {IEEE} {Conference} on {Computer} {Vision}
  and {Pattern} {Recognition}}, pages 762--771, 2018.

\bibitem{revaud_learning_2019}
J.~Revaud, J.~Almazan, R.~S. de~Rezende, and C.~R. de~Souza.
\newblock Learning with {Average} {Precision}: {Training} {Image} {Retrieval}
  with a {Listwise} {Loss}.
\newblock {\em arXiv preprint arXiv:1906.07589}, 2019.

\bibitem{saxena_learning_2006}
A.~Saxena, S.~H. Chung, and A.~Y. Ng.
\newblock Learning depth from single monocular images.
\newblock In {\em Advances in neural information processing systems}, pages
  1161--1168, 2006.

\bibitem{saxena_make3d:_2008}
A.~Saxena, M.~Sun, and A.~Y. Ng.
\newblock Make3d: {Learning} 3d scene structure from a single still image.
\newblock {\em IEEE transactions on pattern analysis and machine intelligence},
  31(5):824--840, 2008.

\bibitem{tateno_cnn-slam_2017}
K.~Tateno, F.~Tombari, I.~Laina, and N.~Navab.
\newblock Cnn-slam: {Real}-time dense monocular slam with learned depth
  prediction.
\newblock In {\em Proceedings of the {IEEE} {Conference} on {Computer} {Vision}
  and {Pattern} {Recognition}}, pages 6243--6252, 2017.

\bibitem{tsai_frank:_2007}
M.-F. Tsai, T.-Y. Liu, T.~Qin, H.-H. Chen, and W.-Y. Ma.
\newblock {FRank}: a ranking method with fidelity loss.
\newblock In {\em Proceedings of the 30th annual international {ACM} {SIGIR}
  conference on {Research} and development in information retrieval}, pages
  383--390. ACM, 2007.

\bibitem{wang_learning_2018}
C.~Wang, J.~Miguel~Buenaposada, R.~Zhu, and S.~Lucey.
\newblock Learning depth from monocular videos using direct methods.
\newblock In {\em Proceedings of the {IEEE} {Conference} on {Computer} {Vision}
  and {Pattern} {Recognition}}, pages 2022--2030, 2018.

\bibitem{wang_towards_2015}
P.~Wang, X.~Shen, Z.~Lin, S.~Cohen, B.~Price, and A.~L. Yuille.
\newblock Towards unified depth and semantic prediction from a single image.
\newblock In {\em Proceedings of the {IEEE} {Conference} on {Computer} {Vision}
  and {Pattern} {Recognition}}, pages 2800--2809, 2015.

\bibitem{xia_listwise_2008}
F.~Xia, T.-Y. Liu, J.~Wang, W.~Zhang, and H.~Li.
\newblock Listwise approach to learning to rank: theory and algorithm.
\newblock In {\em Proceedings of the 25th international conference on {Machine}
  learning}, pages 1192--1199. ACM, 2008.

\bibitem{xian_monocular_2018}
K.~Xian, C.~Shen, Z.~Cao, H.~Lu, Y.~Xiao, R.~Li, and Z.~Luo.
\newblock Monocular relative depth perception with web stereo data supervision.
\newblock In {\em Proceedings of the {IEEE} {Conference} on {Computer} {Vision}
  and {Pattern} {Recognition}}, pages 311--320, 2018.

\bibitem{xian2020structure}
K.~Xian, J.~Zhang, O.~Wang, L.~Mai, Z.~Lin, and Z.~Cao.
\newblock Structure-guided ranking loss for single image depth prediction.
\newblock In {\em Proceedings of the IEEE/CVF Conference on Computer Vision and
  Pattern Recognition}, pages 611--620, 2020.

\bibitem{xu_pad-net_2018}
D.~Xu, W.~Ouyang, X.~Wang, and N.~Sebe.
\newblock Pad-net: {Multi}-tasks guided prediction-and-distillation network for
  simultaneous depth estimation and scene parsing.
\newblock In {\em Proceedings of the {IEEE} {Conference} on {Computer} {Vision}
  and {Pattern} {Recognition}}, pages 675--684, 2018.

\bibitem{zhang_joint_2018}
Z.~Zhang, Z.~Cui, C.~Xu, Z.~Jie, X.~Li, and J.~Yang.
\newblock Joint task-recursive learning for semantic segmentation and depth
  estimation.
\newblock In {\em Proceedings of the {European} {Conference} on {Computer}
  {Vision} ({ECCV})}, pages 235--251, 2018.

\bibitem{zhang_pattern-affinitive_2019}
Z.~Zhang, Z.~Cui, C.~Xu, Y.~Yan, N.~Sebe, and J.~Yang.
\newblock Pattern-{Affinitive} {Propagation} across {Depth}, {Surface} {Normal}
  and {Semantic} {Segmentation}.
\newblock In {\em Proceedings of the {IEEE} {Conference} on {Computer} {Vision}
  and {Pattern} {Recognition}}, pages 4106--4115, 2019.

\bibitem{zhou_unsupervised_2017}
T.~Zhou, M.~Brown, N.~Snavely, and D.~G. Lowe.
\newblock Unsupervised learning of depth and ego-motion from video.
\newblock In {\em Proceedings of the {IEEE} {Conference} on {Computer} {Vision}
  and {Pattern} {Recognition}}, pages 1851--1858, 2017.

\bibitem{zoran_learning_2015}
D.~Zoran, P.~Isola, D.~Krishnan, and W.~T. Freeman.
\newblock Learning ordinal relationships for mid-level vision.
\newblock In {\em Proceedings of the {IEEE} {International} {Conference} on
  {Computer} {Vision}}, pages 388--396, 2015.

\end{thebibliography}
}

\end{document}